\documentclass[10pt,letterpaper]{article}
\usepackage{bbm}
\usepackage{amssymb}
\usepackage{multirow}
\usepackage{subfigure}
\usepackage{url}
\usepackage{slashbox}
\usepackage{hyperref}
\usepackage{booktabs}
\usepackage[dvips]{graphicx}
\usepackage{latexsym,amsmath,amssymb}
\usepackage{geometry} 
\date{}

\begin{document}
\title{Single image super-resolution by approximated Heaviside functions}
\author{
Liang-Jian Deng\thanks{School of Mathematical Sciences, University of Electronic Science and Technology of China, Chengdu, Sichuan, 611731, P. R. China. ({\tt liangjian1987112@126.com}).}
\and Weihong Guo\thanks{The Corresponding Author. Department of Mathematics, Case Western Reserve University, Cleveland, OH, 44106, USA. ({\tt wxg49@case.edu}).}
\and Ting-Zhu Huang\thanks{School of Mathematical Sciences, University of Electronic Science and Technology of China, Chengdu, Sichuan, 611731, P. R. China. ({\tt tingzhuhuang@126.com}).}
        }

\maketitle

\begin{abstract}
Image super-resolution is a process to enhance image resolution. It is widely used in medical imaging, satellite imaging, target recognition, etc. In this paper, we conduct continuous modeling and assume that the unknown image intensity function is defined on a continuous domain and belongs to a space with a redundant basis. We propose a new iterative model for single image super-resolution based on an observation: an image is consisted of smooth components and non-smooth components, and we use two classes of approximated Heaviside functions (AHFs) to represent them respectively. Due to sparsity of the non-smooth components, a $L_{1}$ model is employed. In addition, we apply the proposed iterative model to image patches to reduce computation and storage. Comparisons with some existing competitive methods show the effectiveness of the proposed method.

\end{abstract}

\section{Introduction}
Image super-resolution (SR) is to estimate a high-resolution (HR) image from one or multiple low-resolution (LR) images. When there is only \emph{one} low-resolution input, it is called \emph{single} image super-resolution. The other case is called \emph{multiple} image super-resolution. Compared to multiple image super-resolution, single image super-resolution is more practical when only one low-resolution image is available. Obviously, it is more challenging. In this paper, we address single image super-resolution. We can only collect low-resolution images sometimes because of the limitation of hardware devices and high cost. For instance, it needs more cost and time  to get high-resolution images in medical imaging MRI. Due to long distance and air turbulence, we can not get high-resolution images for synthetic aperture radar (SAR) and satellite imaging. In target recognition, we may not always get high-resolution videos to make recognition accurately. Thus, it is important and useful to develop a more effective image super-resolution algorithm.

There are mainly four categories of image super-resolution methods: interpolation-based methods, learning-based methods, statistics-based methods and other methods. These methods are however not completely independent. For instance, some of statistics-based methods may involve learning strategies.

Nearest-neighbor interpolation and bicubic interpolation are two classical interpolation methods. Nearest-neighbor interpolation estimates intensity of an unknown location using that of its nearest neighbor point. It however often creates jaggy effect. Bicubic interpolation is to interpolate unknown intensity by utilizing a cubic kernel, and may cause blur effect. More interpolation methods have been proposed recently \cite{LiOrchard2001,MuellerLu2007,ZhangWu2006,ChaLee2014,Getreuer2011,Getreuer2011siam,WangWu2014}. In \cite{Getreuer2011siam}, it introduces a contour stencils method to estimate the image contours based on total variation (TV) along curves. This method can distinguish lines of different orientations, curves, corners, etc., with a computationally efficient formula so that the resulted high-resolution images preserve image details well. In \cite{WangWu2014}, Wang et al. propose a fast image super-resolution method on the displacement field. This method is based on two stages, one stage uses an interpolation technique to pick up the low-frequency image and the other employs a reconstruction method to recover local high-frequency structures.

Recently, many image super-resolution methods are learning-based \cite{FreemanPasztor2000,2FreemanPasztor2000,SunZheng2003,TappenRussell2003,GurZalevsky2007,FreemanJones2002,SunSun2008,KimChoi2009,2KimChoi2009,
Liyakathunisa2008,ZhengBouzerdoum2010,XieChen2010,TaiLiu2010,ChangYeung2004,FernandezCandes2013,YangWright2008,YangWright2010,YangWang2011,ZhaoYang2011,
HeQi2013,DongShi2010,ZeydeElad2012,KimKwon2010,BevilacquaRoumy2012,TimofteDe2013,TimofteDe2014,DongLoy2014}. Learning-based methods need two large training data sets, one formed by low-resolution images and the other by the corresponding high-resolution images. We learn a relation between the two training data sets, and then apply the relation to a given low-resolution image to obtain a high-resolution image. They rely heavily on the selection of training data and involves expensive computation. They are not single image super-resolution approaches in strict sense as they require extra training data. In \cite{YangWright2010}, Yang et al. propose a sparse signal representation method for single image super-resolution.
This method jointly trains two dictionaries for the low-resolution and high-resolution image patches, then applies the generated dictionaries to a low-resolution input to get the corresponding high-resolution output. In \cite{DongLoy2014}, Dong et al. first apply a deep learning method for single image super-resolution. The method learns a mapping between low-resolution and high-resolution images. The mapping is represented by a deep convolutional neural network (CNN) that takes the low-resolution image as the input and outputs the high-resolution image.


Statistics-based methods such as Maximum a Posterior (MAP) and Maximum Likelihood estimator (MLE) are a popular tool for image super-resolution   \cite{FarsiuRobinson2004,CapelZisserman2000,Fattal2007}.  Fattal \cite{Fattal2007} utilizes statistical edge dependency relating edge features in low and high resolution images to preserve sharp edges. In \cite{FarsiuRobinson2004}, an alternative approach using $L_{1}$ norm minimization and a bilateral prior based robust regularization, is proposed by Farsin et al.

Many other methods have also been proposed for image super-resolution. Examples are hybrid method \cite{DanielBagon2009}, pixel classification method \cite{AtkinsBouman1999,AtkinsBouman2001}, frequency technique \cite{BormanStevenson1998}, reconstruction method \cite{IraniPeleg1990,IraniPeleg1993,KomatsuIgarashi1993,ChatterjeeMukherjee2007,TakedaFarsiu2007} and others \cite{ViolaFitzgibbon2012,CandesFernandez,FreedmanFattal2011,ShanLi2008,ChambollePock2011}.

Most existing image super-resolution methods are based on discrete models.
In this paper, we propose a continuous model based on approximated Heaviside functions (AHFs) for single image super-resolution.
Because an image normally consists of smooth components and non-smooth components, we model an image as a sum of two classes of AHFs, one representing smooth components and the other depicting non-smooth components, e.g., step edges. Due to the sparsity of non-smooth components, we design a $L_{1}$ regularization model and solve it by alternating direction method of multipliers (ADMM). An iterative model based on the $L_{1}$ model is proposed to preserve more details. We apply the coefficients, computed by the iterative model, to generate high-resolution images at any upscaling factors. In particular, the iterative model is applied to image patches, aiming to get cheap computation and storage. For images with smooth backgrounds, the proposed method may cause ring artifacts and we develop a strategy utilizing image gradient to reduce them. Numerical experiments show the proposed method is a competitive method comparing with existing excellent methods.

To the best of our knowledge, this is the first work to use AHFs on image super-resolution. The proposed method can preserve not only edges, but also the high-frequency details on non-edge regions. The proposed method is completely single image super-resolution method without using training data.

The organization of this paper is as follows. In Section 2, we review Heaviside function and show why we can use it for image super-resolution. In Section 3, we present the proposed method, including the corresponding new model, new algorithm and other strategies. Visual and quantitative experiments are given in Section 4. Finally, we draw conclusions in Section 5.

\section{Heaviside function}
\begin{figure}
\begin{center}
\includegraphics[width=4in,height=2in]{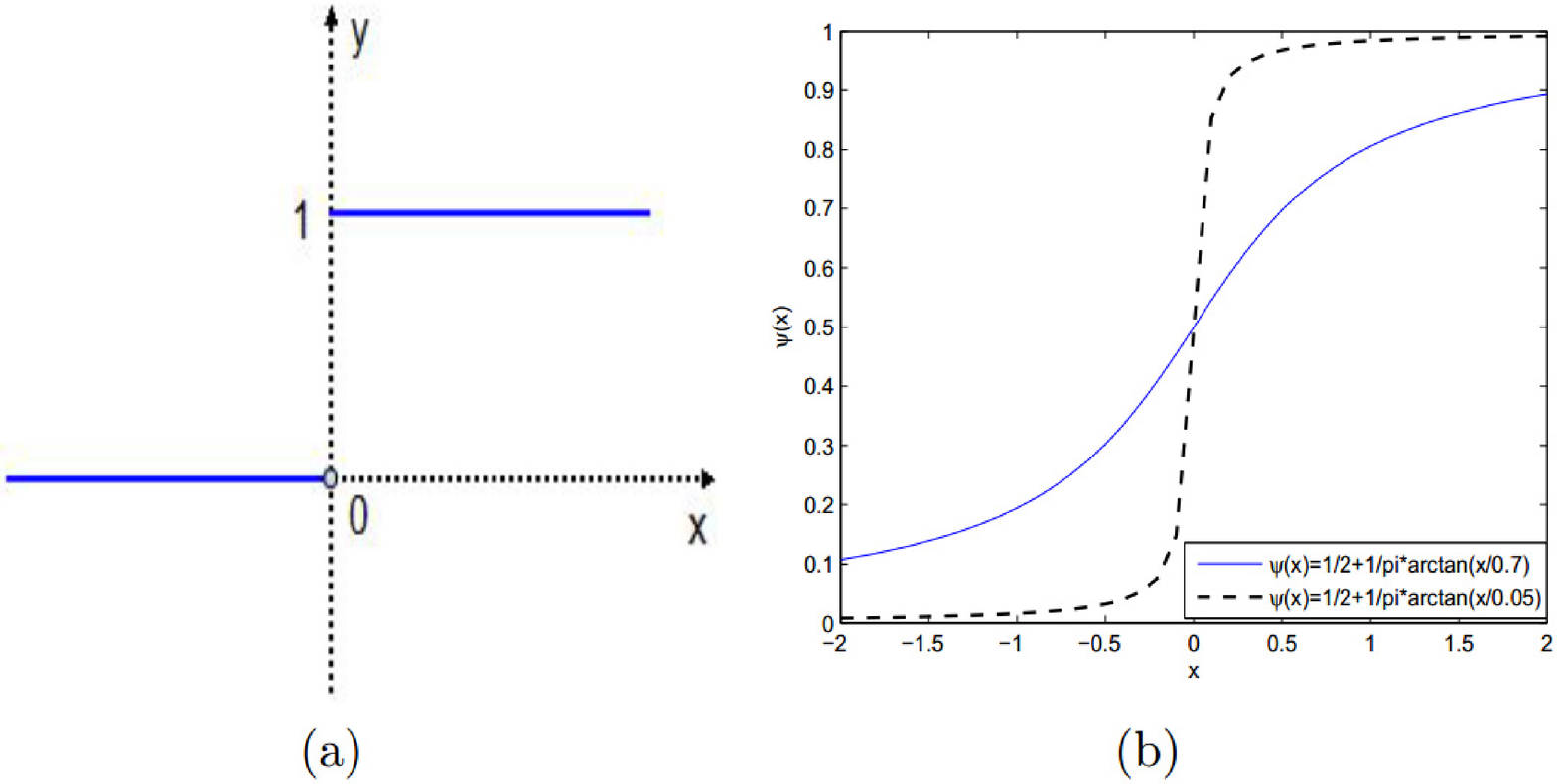}

\end{center}
   \caption{(a) Heaviside function; (b) Two approximated Heaviside functions with $\xi = 0.7$ (blue solid line) and $\xi = 0.05$ (black dash line), the smaller $\xi$ the sharper edge (color images are better visualized in the pdf file).}
\label{fig:1}
\end{figure}

Heaviside function, or Heaviside step function, is defined as follows (see Fig. \ref{fig:1}(a))
\begin{equation}\label{eq:stepheaviside}
\phi(x) = \left\{ {\begin{array}{*{20}c}
   0, \hfill & \indent  x<0, \hfill  \\
    1, \hfill & \indent  x\geq0, \hfill  \\
\end{array}} \right.
\end{equation}
which is singular at $x = 0$. In practice, we usually use its approximation, called approximated Heaviside function (AHF), such as $\frac{1}{1+e^{-2x/\xi}}$ and $\frac{1}{2} + \frac{1}{\pi}\text{arctan}(\frac{x}{\xi})$ which approximate to $\phi(x)$ when $\xi \rightarrow 0$. In this paper, we use the following approximated Heaviside function,
\begin{equation}\label{eq:1dheaviside}
\psi(x) = \frac{1}{2} + \frac{1}{\pi}\text{arctan}(\frac{x}{\xi}),
\end{equation}
where $\xi\in \mathcal{R}$ actually controls the smoothness. The smaller $\xi$, the sharper edge (see Fig. \ref{fig:1}(b)).

From \cite{KainenKurkova2003}, we know that any function in $\mathcal{L}_{p}([0,1]^{d})$, $p\in [1, \infty)$, has a best approximation by linear combinations of $m$ characteristic functions of half-space, and $m$ is any positive integer. Let $H_{d}$ be a set of functions on $[0, 1]^{d}$ defined as
\begin{equation}\label{eq:Hd}
H_{d} = \{f: [0, 1]^{d} \rightarrow \mathcal{R}: f(\textbf{x}) = \psi(\textbf{v}\cdot \textbf{x} + c), \textbf{v}\in \mathcal{R}^{d}, c\in \mathcal{R} \},
\end{equation}
where $\psi$ is an approximated Heaviside function. $H_{d}$ is the set of characteristic function of closed half-spaces of $R^{d}$. 

\textbf{Theorem 1} (see \cite{KainenKurkova2003})
\emph{For any positive integer $d$, define $span_{m}H_{d}$ as $\{\sum_{i=1}^{m}\omega_{i}\psi(\textbf{v}_{i}\cdot \textbf{x} + c_{i})\}$, where $\omega_{i}\in \mathcal{R}$ and $\textbf{v}_{i}\in \mathcal{R}^{d}$ and $c_{i}\in \mathcal{R}$, then it is known that $\text{U}_{m\in \mathcal{N}^{+}}span_{m}H_{d}$ is dense in $(\mathcal{L}_{p}([0, 1]^{d}),\|\cdot\|_{p})$, $p\in [1, \infty)$ }.

\textbf{Theorem 2} (see \cite{KainenKurkova2003})
\emph{For every positive integers $m, d$ and every $p\in [1, \infty)$, $span_{m}H_{d}$ is approximately a compact subset of $(\mathcal{L}_{p}([0, 1]^{d}),\|\cdot\|_{p})$.}

In practical computing, one only can afford to use $span_{m}H_{d}$ for a \emph{finite} $m$.

In our work, we focus on image super-resolution, i.e., $d = 2$. We assume the underlying image intensity function $f$ is defined on $[0, 1]^{2}$ and $f\in \mathcal{L}_{p}([0, 1]^{2})$ with $p\in [1, \infty)$, $f$ can be approximated by the following equation,
\begin{equation}\label{eq:appheaviside}
f(\textbf{z}) = \sum_{j=1}^{m}\omega_{j}\psi(\textbf{v}_{j}\cdot \textbf{z} + c_{j}),
\end{equation}
where $\omega_{j}\in \mathcal{R}$, $\textbf{v}_{j}\in \mathcal{R}^{2}$ and $c_{j}\in \mathcal{R}$. We discreatize $\textbf{v}_{j}=\{(cos\theta_{t}, sin\theta_{t})', t = 1, 2, \cdots, k\}$ to denote $k$ different directions, and $c_{j} =\{\frac{1}{q}, \frac{2}{q}, \frac{3}{q}, \cdots, 1\}$ to denote discrete positions, $m = kq$, $\textbf{z} = (x, y)'$. We fix $q$ as the total number of pixels of the input image. Furthermore, $\{\psi(\textbf{v}_{j}\cdot \textbf{z} + c_{j})\}_{j=1}^{m}$ with a specific $\xi$ is called a class of AHFs. In particular, we can describe edges of different orientations $\theta_{t}$ at the locations $c_{j}$ when setting a small $\xi$. For an image $L\in \mathcal{R}^{n_{1}\times n_{2}}$, we assume it is a discretization of intensity function $f$ on $[0, 1]^{2}$, i.e.,
$L_{i,j}=f(x_{i}, y_{j}), x_{i}=\frac{i}{n_{1}}, y_{j}=\frac{j}{n_{2}}$, $i = 1,2,\cdots, n_{1}$, $j = 1,2,\cdots, n_{2}$. We can rewrite Eq. (\ref{eq:appheaviside}) as matrix-vector form, $L \approx \Psi\omega$, where $\Psi\in \mathcal{R}^{n\times m}, f\in \mathcal{R}^{n}, \omega\in \mathcal{R}^{m}$ with $n = n_{1}n_{2}, m = kq$.
Once we have computed coefficient $\omega$, we can make image super-resolution with upscaling factor $s$ possible by the equation $\widetilde{\Psi}\omega$, where $\widetilde{\Psi}\in \mathcal{R}^{N\times m}$ is obtained similarly with $\Psi$, $N = s^{2}n_{1}n_{2}$.

\begin{figure}
\begin{center}
\includegraphics[width=4.8in,height=2.7in]{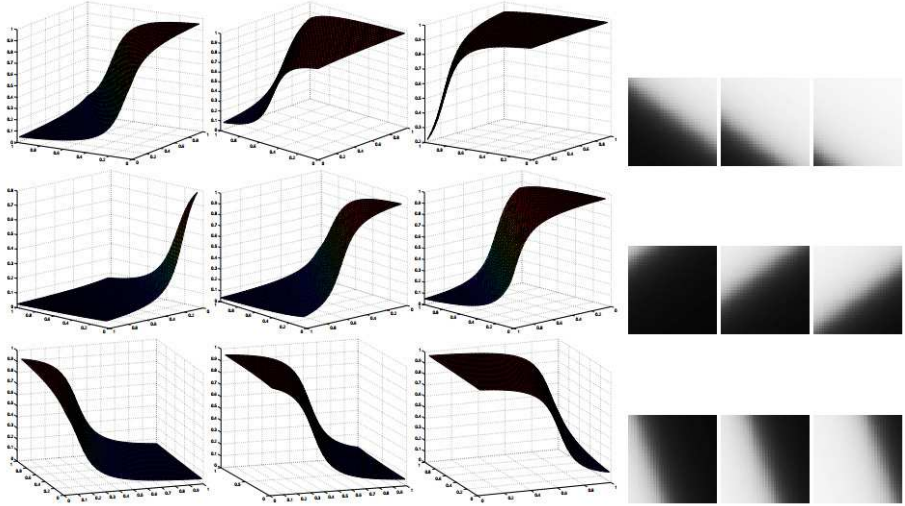}

\end{center}
   \caption{Left panel: surface images of $\psi$ under $\xi = 0.1$ and nine random parameter pairs $(\theta, c)$;  Right panel: the corresponding 2D intensity images. From left to right and then from top to bottom: $(\frac{4\pi}{5}, \frac{51}{1024})$, $(\frac{4\pi}{5}, \frac{25}{64})$, $(\frac{4\pi}{5}, \frac{175}{256})$, $(\frac{6\pi}{5}, \frac{135}{1024})$, $(\frac{6\pi}{5}, \frac{1}{2})$, $(\frac{6\pi}{5}, \frac{25}{32})$, $(\frac{8\pi}{5}, \frac{5}{64})$, $(\frac{8\pi}{5}, \frac{75}{256})$, $(\frac{8\pi}{5}, \frac{75}{128})$ (for better visualization, we make slightly rotation to some surface images).}
\label{fig:p}
\end{figure}

\begin{figure}
\begin{center}
\includegraphics[width=4.8in,height=2.7in]{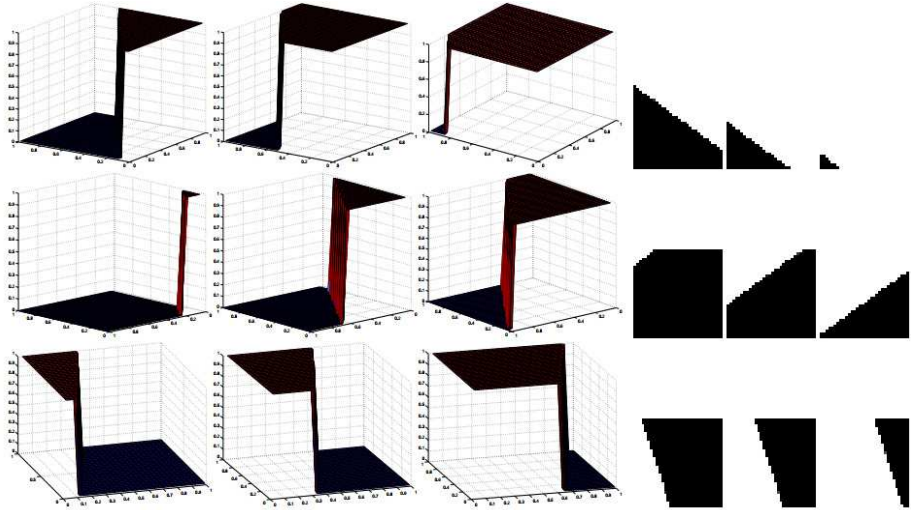}

\end{center}
   \caption{Left panel: surface images of $\psi$ under $\xi = 10^{-4}$ and the same nine parameter pairs $(\theta, c)$ as in Fig. \ref{fig:p}; Right panel: the corresponding 2D intensity images. (for better visualization, we make slightly rotation to some surface images). }
\label{fig:pp}
\end{figure}

\section{The proposed model and its solution}

\subsection{The proposed $L_{1}$ model and its solution}
From Fig. \ref{fig:p} and Fig. \ref{fig:pp}, it is easy to know that the corresponding AHFs with big $\xi$ represent smooth components well, and the corresponding AHFs with small $\xi$ depict non-smooth components such as edges well. An image is generally consisted of smooth components and non-smooth components, thus we represent an image with two classes of AHFs, one with big parameter $\xi_{1}$ to represent smooth components (forming $\Psi_{1}$) and the other with small $\xi_{2}$ to depict non-smooth components (forming $\Psi_{2}$). By this strategy, the vector-form image $L$, still denoted by $I$, can be approximated by the following discrete formula,
\begin{equation}  \label{eq:combineAHF}
L \approx \Psi_{1}\beta_{1} + \Psi_{2}\beta_{2},
\end{equation}
where $\Psi_{1}, \Psi_{2} \in \mathcal{R}^{n\times m}$, and $\beta_{1}$, $\beta_{2} \in \mathcal{R}^{m\times 1}$ are representation coefficients. The following task is to establish a model and compute the representation coefficients $\beta_{1}$ and $\beta_{2}$, and then apply the computed coefficients to get super-resolution images by $\hat{f} = \widetilde{\Psi}_{1}\beta_{1} + \widetilde{\Psi}_{2}\beta_{2}$, where $\widetilde{\Psi}_{1}, \widetilde{\Psi}_{2} \in \mathcal{R}^{N\times m}$ are generated similarly as $\widetilde{\Psi}$ using finer grids.

For an image, smooth components can be mainly represented by $\Psi_{1}\beta_{1}$, non-smooth components such as edges are depicted by $\Psi_{2}\beta_{2}$.  Since non-smooth components are sparse in images. For instance, edges are not everywhere but distribute sparsely in most images. Thus we use $l_{1}$ sparsity on $\beta_{2}$. The optimization model is
\begin{equation}  \label{eq:model}
\min_{\beta_{1},\beta_{2}} \|L - \Psi_{1}\beta_{1} - \Psi_{2}\beta_{2}\|_{2}^{2} + \lambda_{1}\|\beta_{1}\|_{2}^{2} + \lambda_{2}\|\beta_{2}\|_{1},
\end{equation}
where $\lambda_{1}$, $\lambda_{2}$ are regularization parameters, $L$ represents a low-resolution input.

Since $L_{1}$ term is not differentiable, we make a variable substitution for $\beta_{2}$, and rewrite model (\ref{eq:model}) as
\begin{equation}  \label{eq:admm}
\min_{\beta_{1},\beta_{2}} \|L - \Psi_{1}\beta_{1} - \Psi_{2}\beta_{2}\|_{2}^{2} + \lambda_{1}\|\beta_{1}\|_{2}^{2} + \lambda_{2}\|u\|_{1}, \\
\indent \text{s.t.,}\indent u = \beta_{2}, \\
\end{equation}
where $u$ is the substitution variable. In particular, Eq. (\ref{eq:admm}) can be rewritten as
\begin{equation}  \label{eq:sepadmm}
\min_{\beta} \|L - \Psi\beta\|_{2}^{2} + \lambda_{1}\|A\beta\|_{2}^{2} + \lambda_{2}\|u\|_{1}, \\
\indent \text{s.t.,}\indent u = B\beta, \\
\end{equation}
where $\Psi = (\Psi_{1}, \Psi_{2})$, $\beta = (\beta_{1}, \beta_{2})'$, $A = (I, \textbf{0})$ and $B = (\textbf{0}, I)$. The optimization problem  (\ref{eq:sepadmm}) is separable w.r.t $\beta$ and $u$. There are many methods to solve (8). We select alternating direction method of multipliers (ADMM)\cite{GoldsteinOsher2009,WangYang2008,HeTao2011}.

The augmented Lagrangian of Eq. (\ref{eq:sepadmm}) is
\begin{equation}  \label{eq:AL}
\mathcal{L}(\beta, u, b)= \|L - \Psi\beta\|_{2}^{2} + \lambda_{1}\|A\beta\|_{2}^{2} + \lambda_{2}\|u\|_{1} + \frac{\rho}{2}\|u - B\beta + b\|_{2}^{2}, \\
\end{equation}
where $b$ is Lagrangian multiplier with proper size. The problem of minimizing $\mathcal{L}(\beta, u, b)$ is solved by iteratively and alternatively solving the following two subproblems:
\begin{align}
& \beta\text{-subproblem}: \min_{\beta} \|L - \Psi\beta\|_{2}^{2} + \lambda_{1}\|A\beta\|_{2}^{2} + \frac{\rho}{2}\|u - B\beta + b\|_{2}^{2},  \label{eq:bpro} \\
& u\text{-subproblem}: \min_{u} \lambda_{2}\|u\|_{1} + \frac{\rho}{2}\|u - B\beta + b\|_{2}^{2}.  \label{eq:upro}
\end{align}

We get the following ADMM algorithm for problem (\ref{eq:sepadmm}):\\

\noindent
\begin{tabular}{l}
\noindent
\textbf{Algorithm 1}\\
\toprule[1.5pt]
\noindent
\textbf{Input:} Given low-resolution image $L$, $\Psi_{1}$, $\Psi_{2}$, $\lambda_{1}$, $\lambda_{2}$, $\rho$\\
\textbf{Output:} $\beta$\\
\midrule[1.5pt]
1. $k\leftarrow 0$, $\beta^{(k)}\leftarrow 0$, $u^{(k)}\leftarrow 0$, $b^{(k)}\leftarrow 0$                          \\
2. \emph{while not converged do} \\
3. \indent $k\leftarrow k+1$  \\
4. \indent $\beta^{(k)}\leftarrow$ solve  subproblem (\ref{eq:bpro}) for $u = u^{(k-1)}$, $b = b^{(k-1)}$,
\indent\indent\indent\indent\indent\indent\indent\indent\\
5. \indent $u^{(k)}\leftarrow$ solve  subproblem (\ref{eq:upro}) for $\beta = \beta^{(k)}$, $b = b^{(k-1)}$,\\
6. \indent $b^{(k)}\leftarrow b^{(k-1)} + (u^{(k)} - B\beta^{(k)})$.\\
7. \emph{End while}.\\
\bottomrule[1.5pt]
\end{tabular}

The $\beta$-subproblem (\ref{eq:bpro}) can be solved by least squares method:
\begin{equation}
\beta = K^{-1}r,
\end{equation}
where $\beta \in \mathcal{R}^{2m\times 1}$, $K = (\Psi^{T}\Psi+\lambda_{1}A^{T}A+\frac{\rho}{2}B^{T}B) \in \mathcal{R}^{2m\times 2m}$, $r = \frac{\rho}{2}(u+b)+\Psi^{T}L \in \mathcal{R}^{2m\times 1}$.

The $u$-subproblem (\ref{eq:upro}) has a closed form solution for each $u_{i}$ (see \cite{WangYang2008})
\begin{equation}  \label{eq:shrink}
u_{i} =  shrink((B\beta)_{i} - b_{i}, \frac{\lambda_{2}}{\rho}),
\end{equation}
where $shrink(a, b) = sign(a)\text{max}(|a-b|,0)$ and $0.(0/0)=0$ is assumed.

\subsection{The iterative AHF method based on the proposed model}
\begin{figure}
\begin{center}
\includegraphics[width=5.6in,height=1.6in]{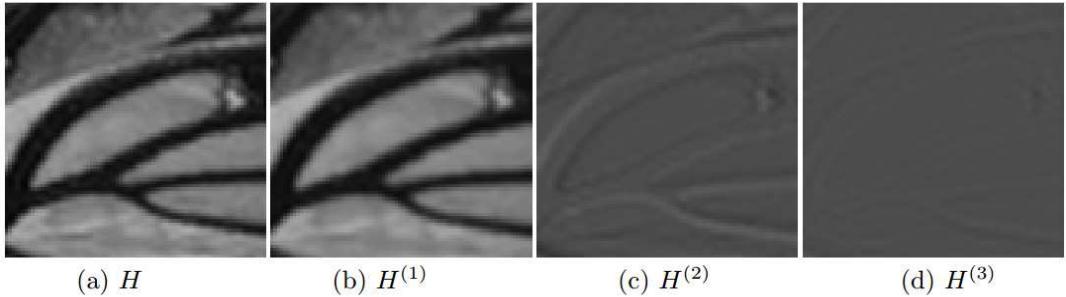}

\end{center}
   \caption{Illustration of step 2 of Algorithm 2 using ``wing'' image; (a) the sum image of $H^{(i)}, i =1, 2, 3$ (we set 3 iterations here) ; (b) the computed image for the first iteration. For better visualization, we add 0.3 to the intensities of $H^{(2)}$ and $H^{(3)}$ to brighten them. From the last two images, we can see that $H^{(2)}$ and $H^{(3)}$ pick up image details.}
\label{fig:iteration}
\end{figure}

Even though model (\ref{eq:model}) takes different behaviors of smooth components and non-smooth components into consideration, we still see some blur in the high-resolution image. We find that the difference image $L - DH^{(1)}$, where $H^{(1)}$ is the resulted high-resolution output by model (\ref{eq:model}) and $D$ is a downsampling operator, contains some residual edges. To pick up more edges and details and thus to make results less blurry, we consider the difference $L - DH^{(1)}$ as a new low-resolution input $L$ of model (\ref{eq:model}) to recompute a residual high-resolution image $H^{(2)}$. This process is repeated until the residual edges are small enough. The sum of $H^{(1)}$ and its residual high-resolution images is the resulted super-resolution image. This iterative strategy can recover more image details (see Fig. \ref{fig:iteration}). The following Algorithm 2 is the proposed iterative AHF algorithm for single image super-resolution. We use bicubic downsampling $D$ in our experiments although Algorithm 2 can work for general downsampling.\\



\noindent
\begin{tabular}{l}
\noindent
\textbf{Algorithm 2} (Single image super-resolution via iterative AHF method)\\
\toprule[1.5pt]
\noindent
\textbf{Input:} one low-resolution image (vector-form): $L \in \mathcal{R}^{n\times 1}$, $\lambda_{1}> 0$, $\lambda_{2}> 0$, $s$: upscaling factor.\\
\indent\indent\quad $\tau$: maximum number of iteration.\\
\textbf{Output:} high-resolution image (vector-form): $\widehat{H}\in \mathcal{R}^{N\times 1}$  \\
\midrule[1.5pt]
1. Construct matrices $\Psi_{1}, \Psi_{2}\in \mathcal{R}^{n\times m}$ on coarse grids and $\widetilde{\Psi}_{1}, \widetilde{\Psi}_{2}\in \mathcal{R}^{N\times m}$ on fine grids (see Section 3.1),\\
\indent where $N = s^{2}n$. \\
2. Initialization: $L^{(1)} = L$. \\
\indent \textbf{for k = 1: $\tau$} \\
\indent\indent a. Compute the coefficients: $(\beta_{1}^{(k)}, \beta_{2}^{(k)})= \text{argmin}\|L^{(k)} - \Psi_{1}\beta_{1} - \Psi_{2}\beta_{2}\|_{2}^2 + \lambda_{1}\|\beta_{1}\|_{2}^{2} + \lambda_{2}\|\beta_{2}\|_{1}$.\\
\indent\indent b. Update the high-resolution image: $H^{(k)} = S^{(k)}+E^{(k)}~\text{where}~S^{(k)}=\widetilde{\Psi}_{1}\beta_{1}^{(k)}, E^{(k)}=\widetilde{\Psi}_{2}\beta_{2}^{(k)}$.\\
\indent\indent c. Downsampling $H^{(k)}$ to coarse grid: $\widetilde{L} = DH^{(k)}$.\\
\indent\indent d. Compute residual: $L^{(k+1)} = L^{(k)} - \widetilde{L}$. \\
\indent \textbf{end} \\
3. Assemble the high-resolution outputs: $S=\sum_{i=1}^{\tau}S^{(i)}$, $E=\sum_{i=1}^{\tau}E^{(i)}$.\\
4. Compute the final high-resolution image: \\
\indent\indent\indent\indent\indent\indent\indent\indent\indent\indent   $\widehat{H} = S + conv(E, p)$, \\
\indent where \emph{conv} represents a convolution operator, and $p$ is a Gaussian kernel with small size. \\
\bottomrule[1.5pt]
\end{tabular}

\begin{figure}
\begin{center}
\includegraphics[width=4.2in,height=1.4in]{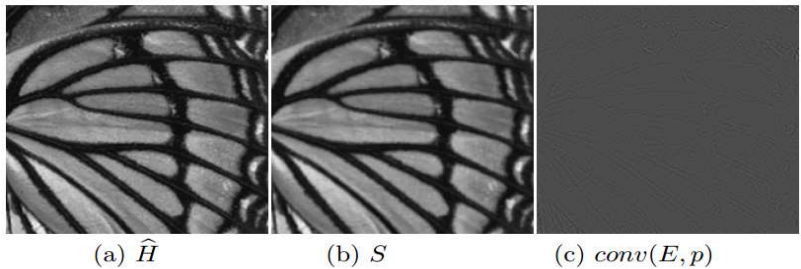}

\end{center}
   \caption{Illustration of step 4 of Algorithm 2 on pathes of a ``wing image; (a) the sum image $S + conv(E, p)$ (see step 4 of Algorithm 2); (b) the image $S$; (c) the image $conv(E, p)$ which contains some edges and high-frequency information away from edges. For better visualization, we add 0.3 to the intensities of $conv(E, p)$ to brighten it.}
\label{fig:smoothedge}
\end{figure}

Note that step 2.a in Algorithm 2 is solved by Algorithm 1. From Fig. \ref{fig:smoothedge}, we know that $E$ represents some edges. In particular, we use a Gaussian kernel $p$ with size $5\times 5$ and standard derivation 1 to make a convolution to reduce the oversharp information on the non-edge regions. In addition, although we introduce some parameters in the algorithm, they are not sensitive and easy to select. We will give a remark on parameter selection in Section 4. This is the proposed iterative method for single image super-resolution.

\subsection{Apply the proposed method to image patches}
We have proposed the iterative AHF method for single image super-resolution, but it can be very expensive if we apply it to a whole image. For instance, if one low-resolution image has size $100\times 100$ and the upscaling factor is 2, then $\Psi_{1}, \Psi_{2}$ with 10 angles should have size $10,000\times 100,000$. It is very expensive to implement Algorithm 2 with big non-sparse matrices. In our work, we apply the iterative AHF method to image patches, so that
we can reduce computation and storage significantly. We set patch size to be $6\times 6$ and overlap to be 2 in our work. Moreover, pixels on the boundary will be handled by bicubic interpolation.

In summary, we use a set of Heaviside functions with sharp edges to represent sharp image edges which can be viewed as a kind of image high-frequency information. In particular, applying the proposed method to each image patches may make image high-frequency information more significant. This is the reason why the proposed method can pick up image high-frequency information.
\section{Numerical experiments}
In this section, two kinds of test images are utilized to illustrate the performance of different methods. One is low-resolution images without high-resolution ground-truth. The other is low-resolution images downsampled from known high-resolution images, for which we can make quantitative comparisons. All experiments are implemented in MATLAB(R2010a) on a laptop of 3.25Gb RAM and Intel(R) Core(TM) i3-2370M CPU: @2.40 GHz.

We compare our method with some competitive image super-resolution methods: nearest-neighbor interpolation, bicubic interpolation, iterative back-projection method (denoted as ``BP'' \cite{IraniPeleg1993}), a edge-directed interpolation (denoted as ``01'TIP'' \cite{LiOrchard2001}), an edge-guided interpolation (denoted as ``06'TIP'' \cite{ZhangWu2006}), a kernel regression method (denoted as ``07'TIP'' \cite{TakedaFarsiu2007}), a fast upsamling method (denoted as ``08'TOG'' \cite{ShanLi2008}), TV-based super-resolution method (denoted as ``11'JMIV'' \cite{ChambollePock2011}), three state-of-the-art learning-based methods (denoted as ``04'CVPR'' \cite{ChangYeung2004}, ``10'TIP'' \cite{YangWright2010} and ``12'BMVC'' \cite{BevilacquaRoumy2012}) and three state-of-the-art interpolation methods (denoted as ``11'IPOL'' \cite{Getreuer2011}, ``11'SIAM'' \cite{Getreuer2011siam} and ``14'TIP'' \cite{WangWu2014}).

For gray images, we can apply the proposed Algorithm 2 to them directly.
For color images such as RGB, we transform test images in RGB color space to ``Ycbcr'' color space which is very popular in image/video processing. ``Y'' represents luma component, ``Cb'' and ``Cr'' are blur-difference and red-difference components, respectively. We only apply the proposed algorithm to the illuminance channel because humans are more sensitive to luminance changes. In addition, we interpolate the color layers (Cb, Cr) using bicubic interpolation. After getting the upsampled images in ``Ycbcr'' color space, we transform them back to the original RGB color space for visualization.

We use root-mean-square error (RMSE) on illuminance channel to evaluate numerical performance,
\begin{equation}  \label{eq:rmse}
\text{RMSE} = \sqrt{\frac{1}{N}\sum_{i=1}^{N}(h_{i} - \widehat{h}_{i})^{2}},
\end{equation}
where $h, \widehat{h}$ are the vector-form of ground-truth image and resulted super-resolution image, respectively. Note that although RMSE is a common numerical criterion, smaller RMSE can not always promise better visual results.

\noindent\textbf{Remark on parameter selection:} For the parameters in Algorithm 1 and Algorithm 2, we set $\rho = 10^{-4}$, $\lambda_{1} = 10^{-2}$, $\lambda_{2} = 10^{-6}$, $\tau = 3$. Matrices $\Psi_{1}, \widetilde{\Psi}_{1}$ with $\xi = 10^{-1}$ and $\Psi_{2}, \widetilde{\Psi}_{2}$ with $\xi = 10^{-4}$. We set 12 angles evenly distributed on $[0, 2\pi]$, i.e., $\theta = \{0, \frac{\pi}{6}, \frac{\pi}{3}, \frac{\pi}{2}, \frac{2\pi}{3}, \frac{5\pi}{6}, \pi, \frac{7\pi}{6}, \frac{4\pi}{3}, \frac{3\pi}{2}, \frac{5\pi}{3}, \frac{11\pi}{6}\}$. We fix $q$ as the total number of pixels of the low-resolution patch, i.e., $q = 36$ for $6\times 6$ patches. Note that, fine tuning of parameters for some images may get better results. However, we unify the parameter selection to illustrate the stability of the proposed method.

\subsection{Super-resolution for generic images}
\begin{figure}
\begin{center}
\includegraphics[width=5.6in,height=2.46in]{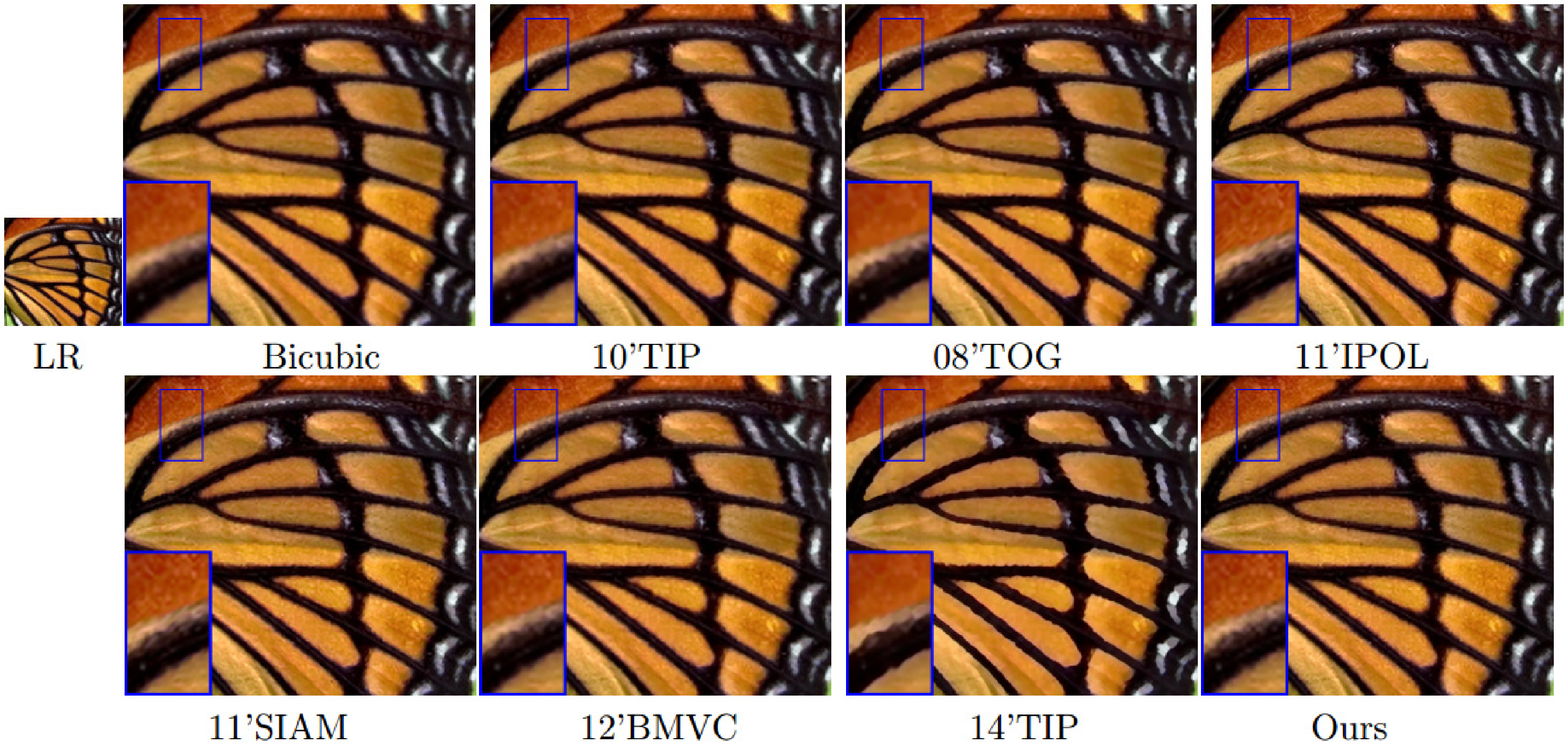}

\includegraphics[width=5.6in,height=2.46in]{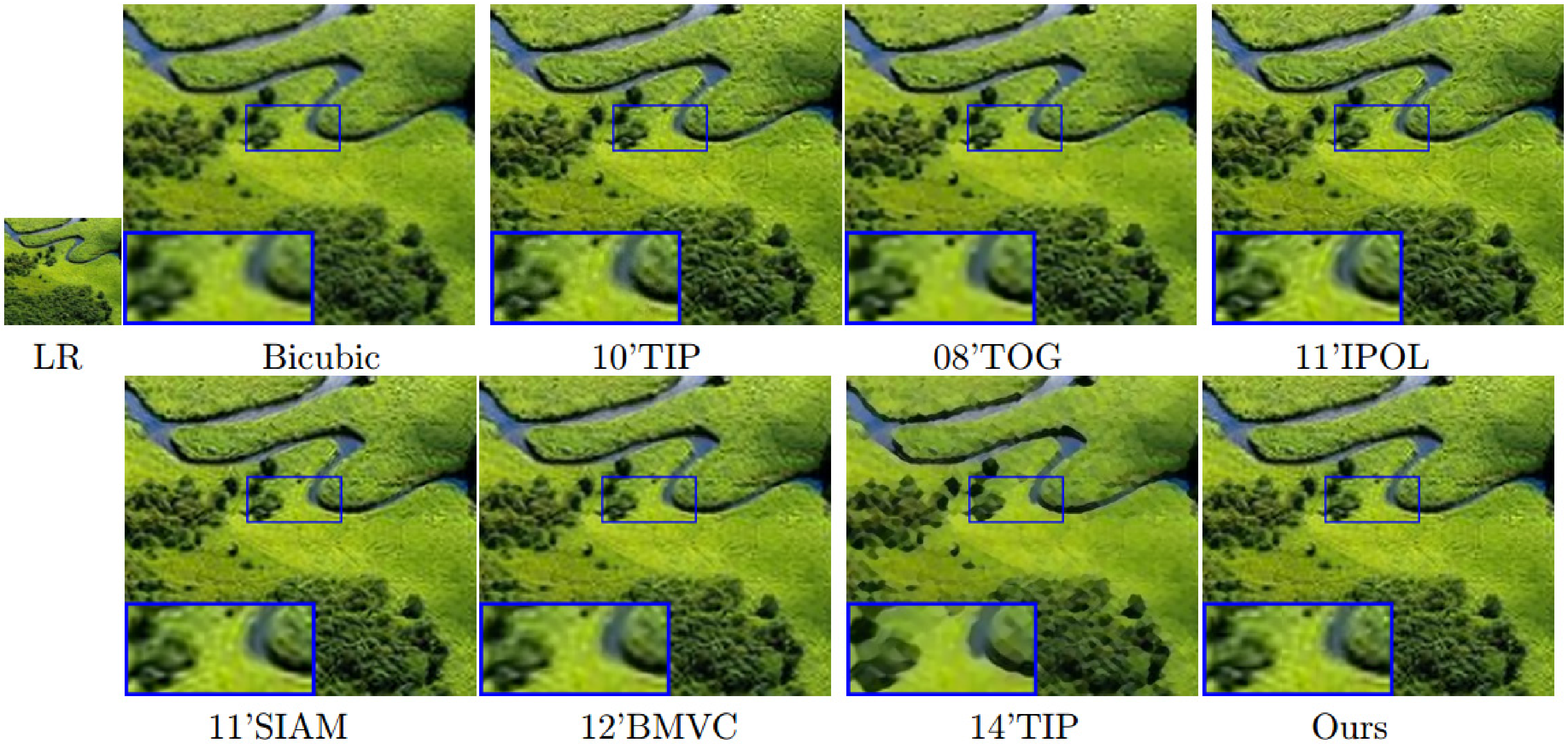}

\end{center}
   \caption{Results of ``butterflywing'' (first row) and ``landscape'' (second row), upscaling factors are all 3. From left to right: low-resolution images, results of bicubic interpolation, ``10'TIP'' \cite{YangWright2010}, ``08'TOG'' \cite{ShanLi2008}, ``11'IPOL''  \cite{Getreuer2011}, ``11'SIAM'' \cite{Getreuer2011siam}, ``12'BMVC'' \cite{BevilacquaRoumy2012}, ``14'TIP'' \cite{WangWu2014} and the proposed method (color images are better visualized in the pdf file).}
\label{fig:wing-ww}
\end{figure}

From Fig. \ref{fig:wing-ww}, we test low-resolution images without known high-resolution images. We compare the proposed method with nearest-neighbor interpolation, bicubic interpolation, two state-of-the-art learning-based methods (``10'TIP'' \cite{YangWright2010} and ``12'BMVC'' \cite{BevilacquaRoumy2012}),  a fast upsampling method (``08'TOG'' \cite{ShanLi2008}) and three state-of-the-art interpolation-based methods (``11'IPOL'' \cite{Getreuer2011}, ``11'SIAM'' \cite{Getreuer2011siam} and ``14'TIP'' \cite{WangWu2014}). The upscaling factors are all 3. From the resulted images, super-resolution images by bicubic interpolation generate blur effect. ``08'TOG'' performs well on image edges, but it smooths the details on the non-edge regions. The two learning-based methods (``10'TIP'' and ``12'BMVC'') perform comparable visually. In addition, ``11'IPOL'' and ``11'SIAM'' also show similar performance. In particular, ``14'TIP'' shows better property for image edges, but it smooths the image details on the non-edge regions. The proposed method outperforms the other methods, especially for image details on the non-edge regions.

\begin{figure}
\begin{center}

\includegraphics[width=5.6in,height=2.6in]{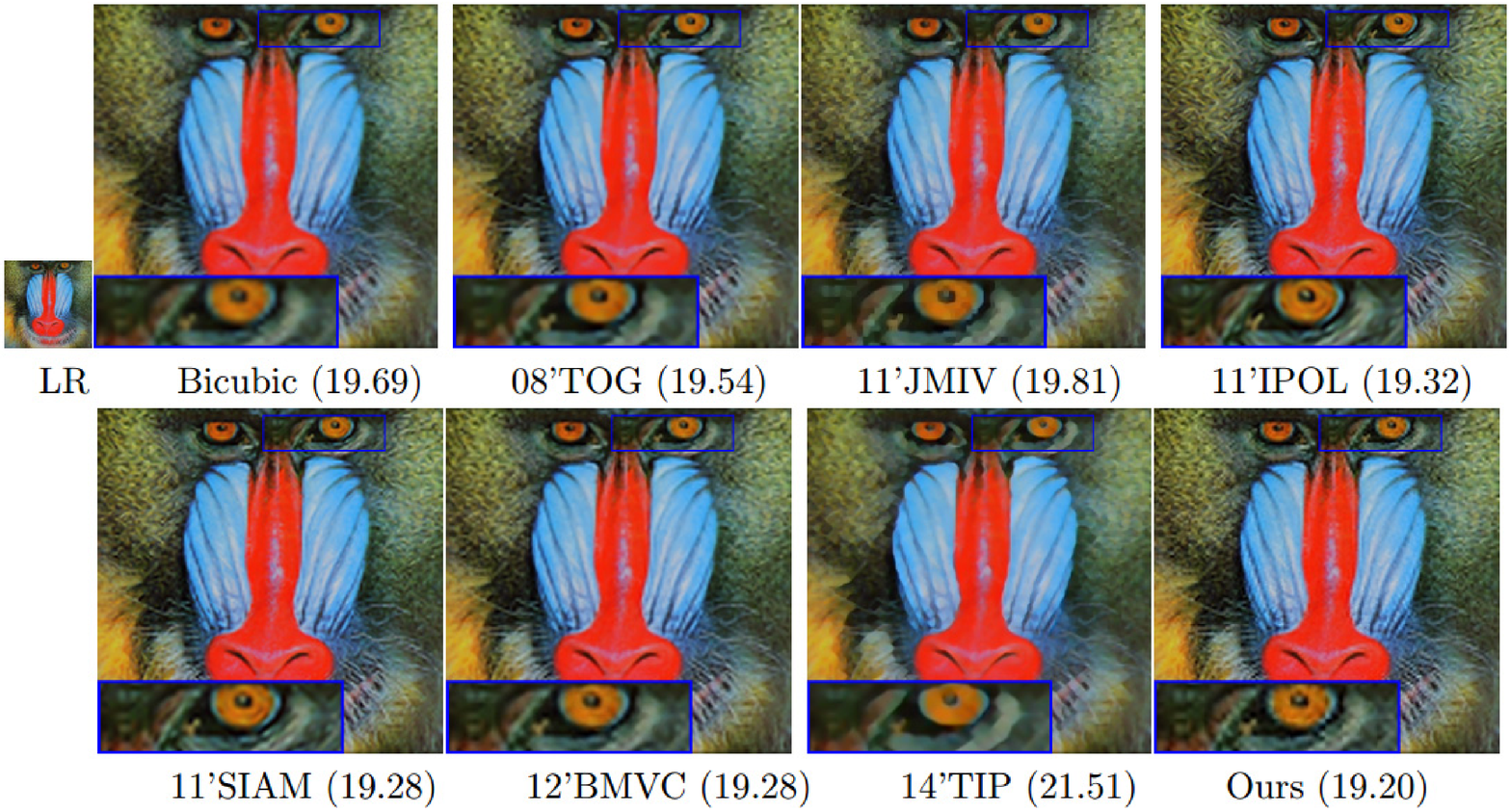}

\includegraphics[width=5.6in,height=2.6in]{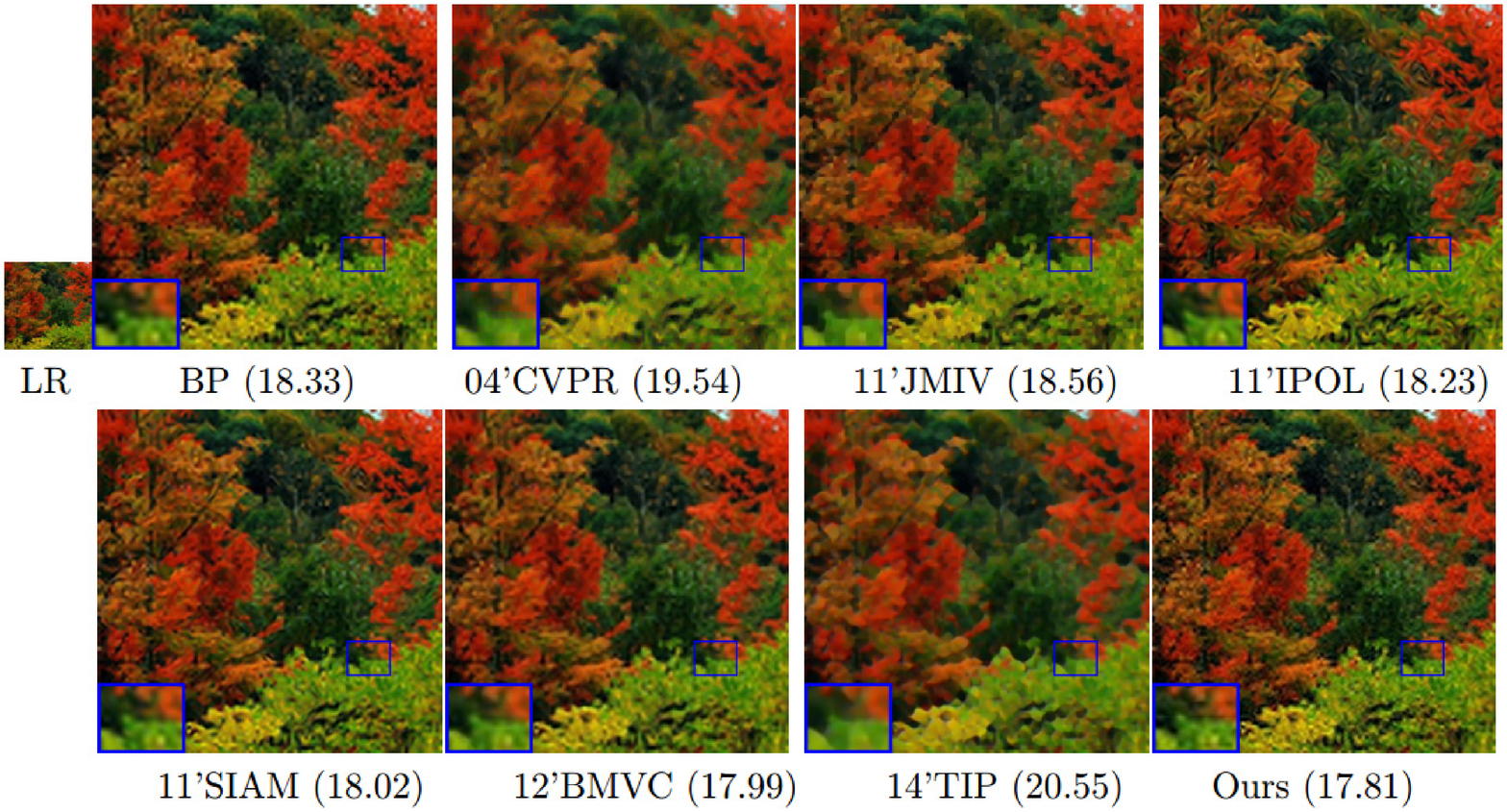}

\end{center}
   \caption{Visual results and the corresponding RMSE of ``baboon'' and ``forest'' by upscaling factors of 4. Compared methods: bicubic interpolation, ``BP'' \cite{IraniPeleg1993}, ``04'CVPR'' \cite{ChangYeung2004}, ``08'TOG'' \cite{ShanLi2008}, ``11'JMIV'' \cite{ChambollePock2011}, ``11'IPOL'' \cite{Getreuer2011}, ``11'SIAM'' \cite{Getreuer2011siam}, ``12'BMVC'' \cite{BevilacquaRoumy2012}, ``14'TIP'' \cite{WangWu2014}, and the proposed method. }
\label{fig:baboon-forest}
\end{figure}

\begin{figure}
\begin{center}

\includegraphics[width=5.6in,height=3.15in]{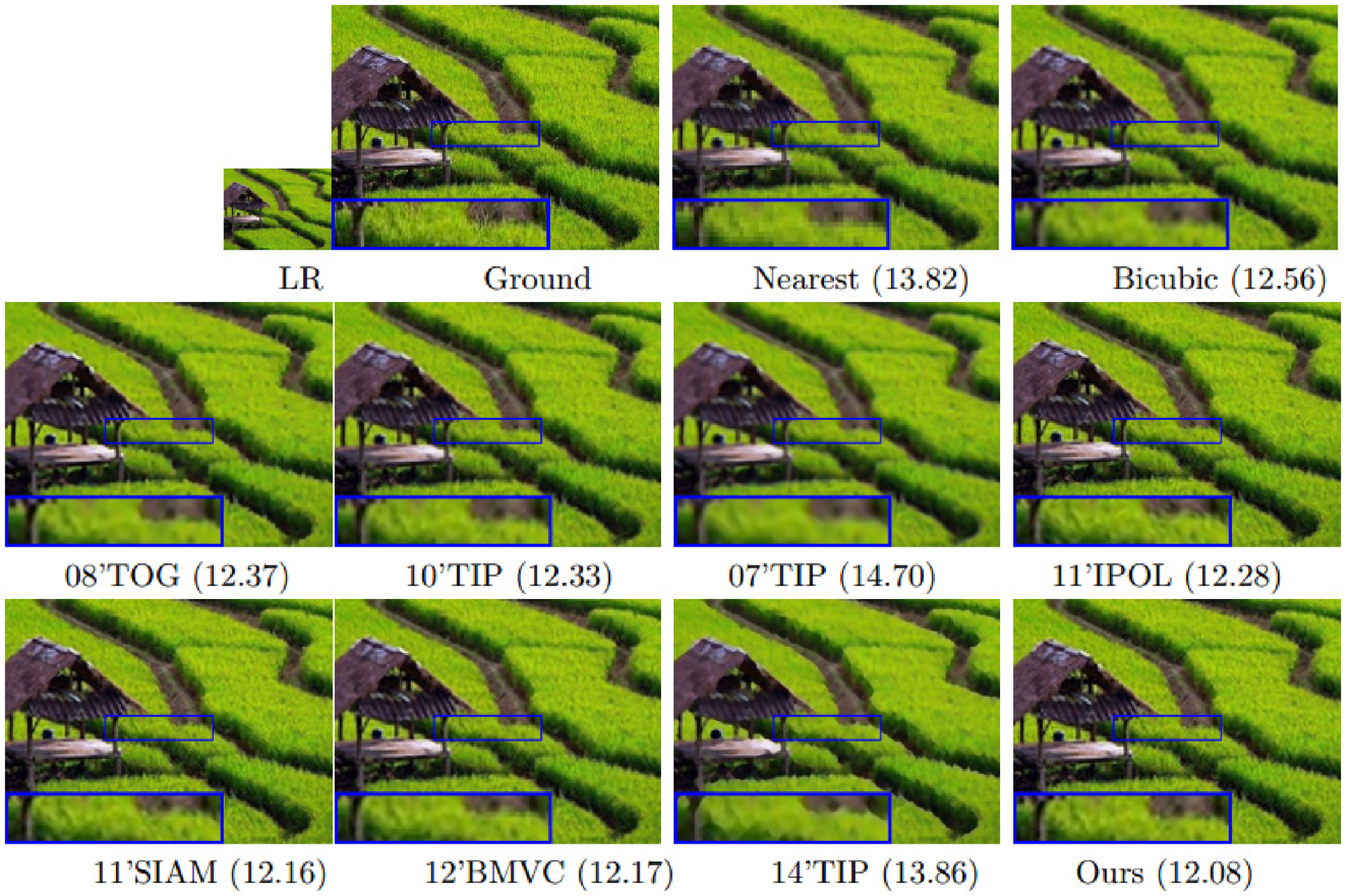}

\includegraphics[width=5.6in,height=2.3in]{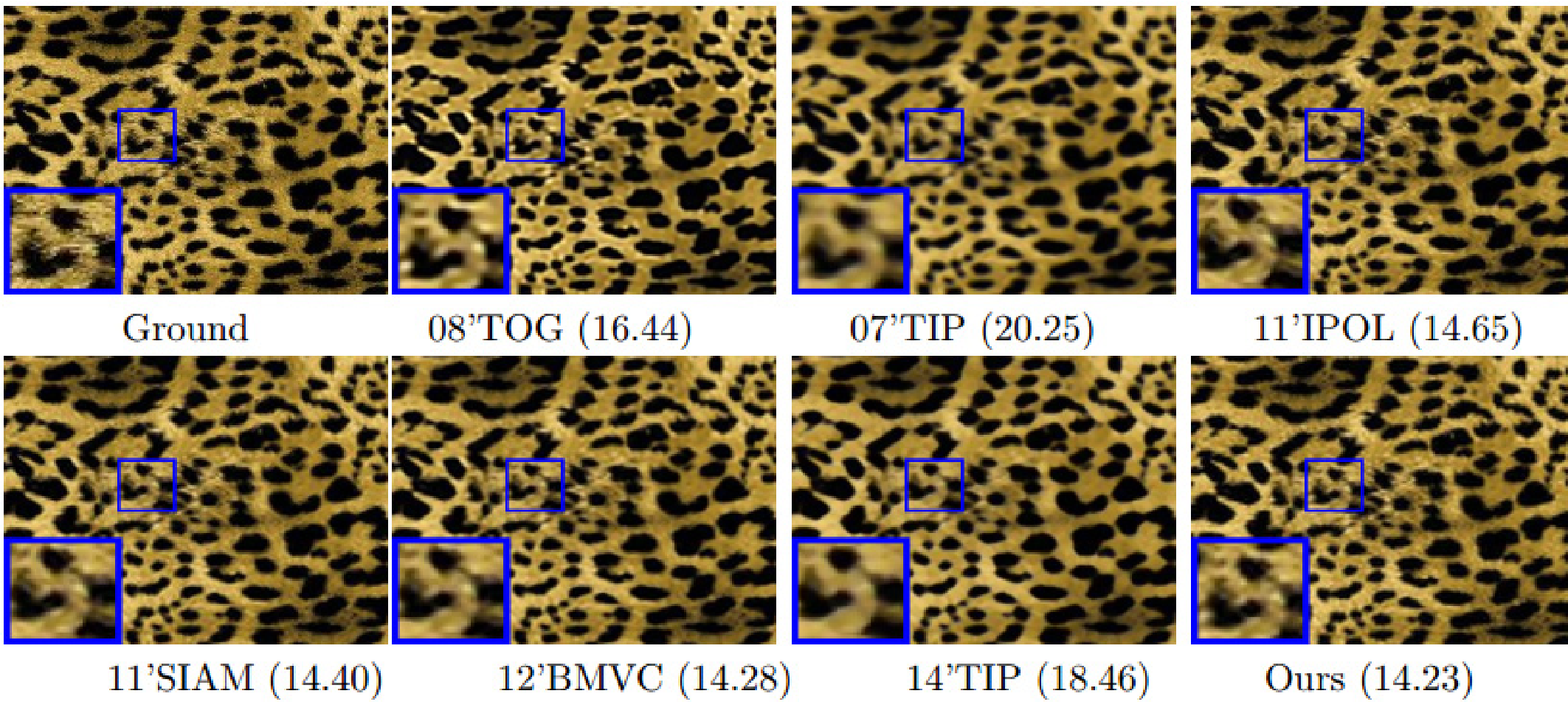}

\end{center}
   \caption{Visual results of ``field'' (upscaling factor 3) and ``leopard'' (upscaling factor 2) and the corresponding RMSE. Compared methods: nearest-neighbor interpolation, bicubic interpolation, ``07'TIP''  \cite{TakedaFarsiu2007}, ``08'TOG'' \cite{ShanLi2008}, ``10'TIP'' \cite{YangWright2010}, ``11'IPOL'' \cite{Getreuer2011}, ``11'SIAM'' \cite{Getreuer2011siam}, ``12'BMVC'' \cite{BevilacquaRoumy2012}, ``14'TIP'' \cite{WangWu2014}, and the proposed method.}
\label{fig:filed-baowen}
\end{figure}

In Fig. \ref{fig:baboon-forest} and Fig. \ref{fig:filed-baowen}, low-resolution test images are generated by downsampling known high-resolution via bicubic interpolation. Because ground-truth images are known, we can evaluate quantitative results using RMSE. The upscaling factors are set to be 4 for ``baboon'' and ``forest'', 3 for ``field'' and 2 for ``leopard''. From these figures, we know that the results of ``08'TOG'' generate sharp edges but smooth out details on non-edge regions. TV-based method ``11'JMIV'' \cite{ChambollePock2011} also obtains sharp edges but has oil-painting effect. Results of iterative back-projection ``BP'' \cite{IraniPeleg1993}, the learning-based method (neighborhood embedding ``04'CVPR'' \cite{ChangYeung2004}) and the kernel regression method ``07'TIP'' \cite{TakedaFarsiu2007} show significant blur. Because the proposed method can be viewed as an interpolation method, we compare our method with three state-of-the-art interpolation methods (``11'IPOL'' \cite{Getreuer2011}, ``11'SIAM'' \cite{Getreuer2011siam} and ``14'TIP'' \cite{WangWu2014}). In addition, we also compare the proposed method with two modern learning-based methods (``10'TIP'' \cite{YangWright2010} and ``12'BMVC'' \cite{BevilacquaRoumy2012}). From Fig. \ref{fig:baboon-forest} and Fig. \ref{fig:filed-baowen}, the proposed method obtains smaller RMSE than the three state-of-the-art interpolation methods and the two learning-based methods, and generates high-resolution images with more image details. Actually, learning-based methods need extra training data that can be viewed as extra prior information while our method does not require any extra data. Note that the interpolation-based method ``14'TIP'' \cite{WangWu2014} keeps very sharp image edges and runs quite fast, but it obtains bad RMSE and smooths image details significantly on the non-edge regions (see the close-ups).  Furthermore, more quantitative results can be found in Tab. \ref{tab:1}. It also demonstrates the effectiveness of the proposed method.


\begin{table*}[!htp]
\caption{RMSE of more test examples}\label{tab:1}
\begin{center}
\begin{tabular}{cccccc}
\toprule[2pt]

   Example(factor)   & Bicubic & 07'TIP \cite{TakedaFarsiu2007}  & 08'TOG \cite{ShanLi2008} & 10'TIP \cite{YangWright2010} & Ours   \\
\midrule[1pt]
  lena(2)    & 5.10 & 9.91 &  6.98 & 4.09 &  \textbf{3.84} \\
\midrule[1pt]
  landscape(2)    & 10.75 & 13.55 &  11.74 & 10.09 &  \textbf{9.63} \\
\midrule[1pt]
  purplebutterfly(2)    & 13.03 & 16.85 &  14.19 & 11.69 &  \textbf{11.43} \\
\midrule[1pt]
     leaf(3)    & 17.51 & 18.23 &  17.55 & 17.39 &  \textbf{17.28} \\
\midrule[1pt]
     dog(3)    & 2.12  & 3.76  &  1.95  & 1.93  &  \textbf{1.85} \\
\midrule[1pt]
     fruits(3)    & 4.13  &  9.09 &  3.55  & 3.75  &  \textbf{3.53} \\
\midrule[1pt]
     babyface(3)    & 5.12  &  8.44 &  4.86  & 4.87  &  \textbf{4.74} \\
\midrule[1pt]
      train(4)    & 12.98 &  15.85 &  12.64 & 12.85 &  \textbf{12.62} \\
\bottomrule[2pt]
\end{tabular}
\end{center}
\end{table*}

\subsection{Super-resolution for the images with smooth backgrounds}
\begin{figure}
\begin{center}
\includegraphics[width=5.0in,height=1.76in]{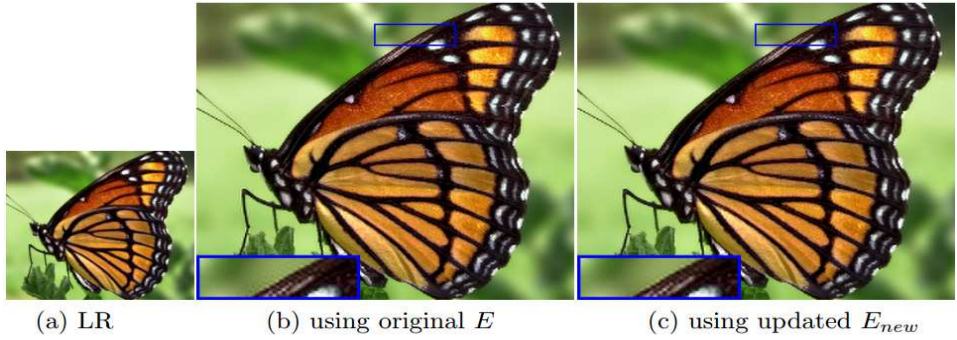}

\end{center}
   \caption{Results of ``butterfly'' by an upscaling factor of 2. Image ``butterfly'' is an image with the smooth background; (a) low-resolution image; (b) super-resolution image by the proposed method using original $E$; (c) super-resolution image by the proposed method using updated $E_{new}$.}
\label{fig:butter}
\end{figure}

The proposed method performs well for natural images. It is a better method than the compared methods. However, the proposed method encounters a drawback: ring artifacts along the large scale edges of images which have smooth backgrounds (see Fig. \ref{fig:butter}(b)). The images with smooth backgrounds are generally obtained from single lens reflex (SLR) camera or other specific scenes. The ring artifacts mainly come from the added non-smooth components (see step 4 of Algorithm 2). In this section, we utilize a method without ring artifacts to make a mask, aiming to discard the ring artifacts of non-smooth components $E$. We learn that bicubic interpolation will not cause ring artifacts (see the resulted images in section 4.1) and run very fast. Thus we use bicubic interpolation as our intermediate method to make the mask. Actually, we only need to update the $E$ in the step 4 of Algorithm 2 by $E_{new}$ which is obtained from the following equations, then we can reduce the ring artifacts significantly.
\begin{equation}\label{eq:Enew}
E_{new} = \text{Mask}.*E,
\end{equation}
where
\begin{equation}  \label{eq:mask}
\text{Mask} = \left\{ {\begin{array}{*{20}c}
   0, \hfill & \indent {\rm if} ~ 0\leq G_{i,j}\leq t, \hfill  \\
    1, \hfill & \indent {\rm otherwise}, \hfill  \\
\end{array}} \right.
\end{equation}
where $G_{i,j}$ is the vector norm of gradient at the location $(i, j)$ of image $B$; $B = Bicubic(L, s)$ where ``\emph{Bicubic}'' represents implementing bicubic interpolation to a low-resolution image and $L$ is a low-resolution image, $s$ is an upscaling factor; notation ``$.*$'' stands for dot product;  $t$ is a thresholding value, and we set $t = 0.05$ in our work. Obviously, this strategy reduces the ring artifacts of super-resolution images (see Fig. \ref{fig:butter}(c)).

\begin{figure}
\begin{center}

\includegraphics[width=5in,height=3in]{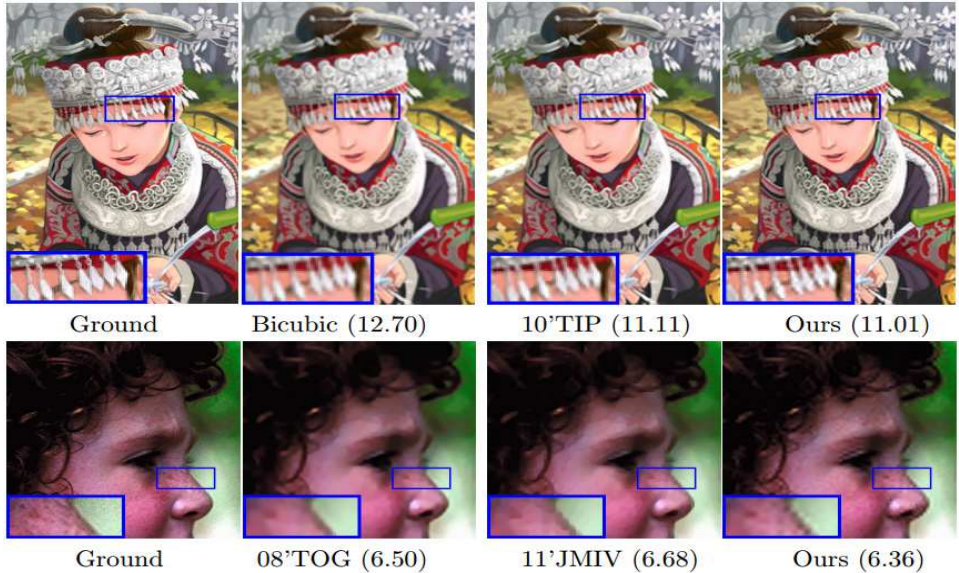}

\end{center}
   \caption{Results of ``comic'' (first row, upscaling factor 2) and ``face'' (second row, upscaling factor 4) and the corresponding RMSE. First row: ground-truth, bicubic interpolation, the learning-based method ``10'TIP'' \cite{YangWright2010} and the proposed method; Second row: ground-truth, the fast upsampling method ``08'TOG''  \cite{ShanLi2008}, TV-based method ``11'JMIV''  \cite{ChambollePock2011} and the proposed. Our method gets smallest RMSE and better visual results.}
\label{fig:comic-face}
\end{figure}

In Fig. \ref{fig:comic-face}, we apply the proposed method to images ``comic'' and ``face'' which have smooth backgrounds. The upscaling factors are set to be 2 and 4, respectively. We can find more details from the close-ups in Fig. \ref{fig:comic-face}. From the figure, the proposed method reduces ring artifacts significantly and performs best, both visually and quantitatively.

\begin{figure}
\begin{center}
\includegraphics[width=5.1in,height=3.84in]{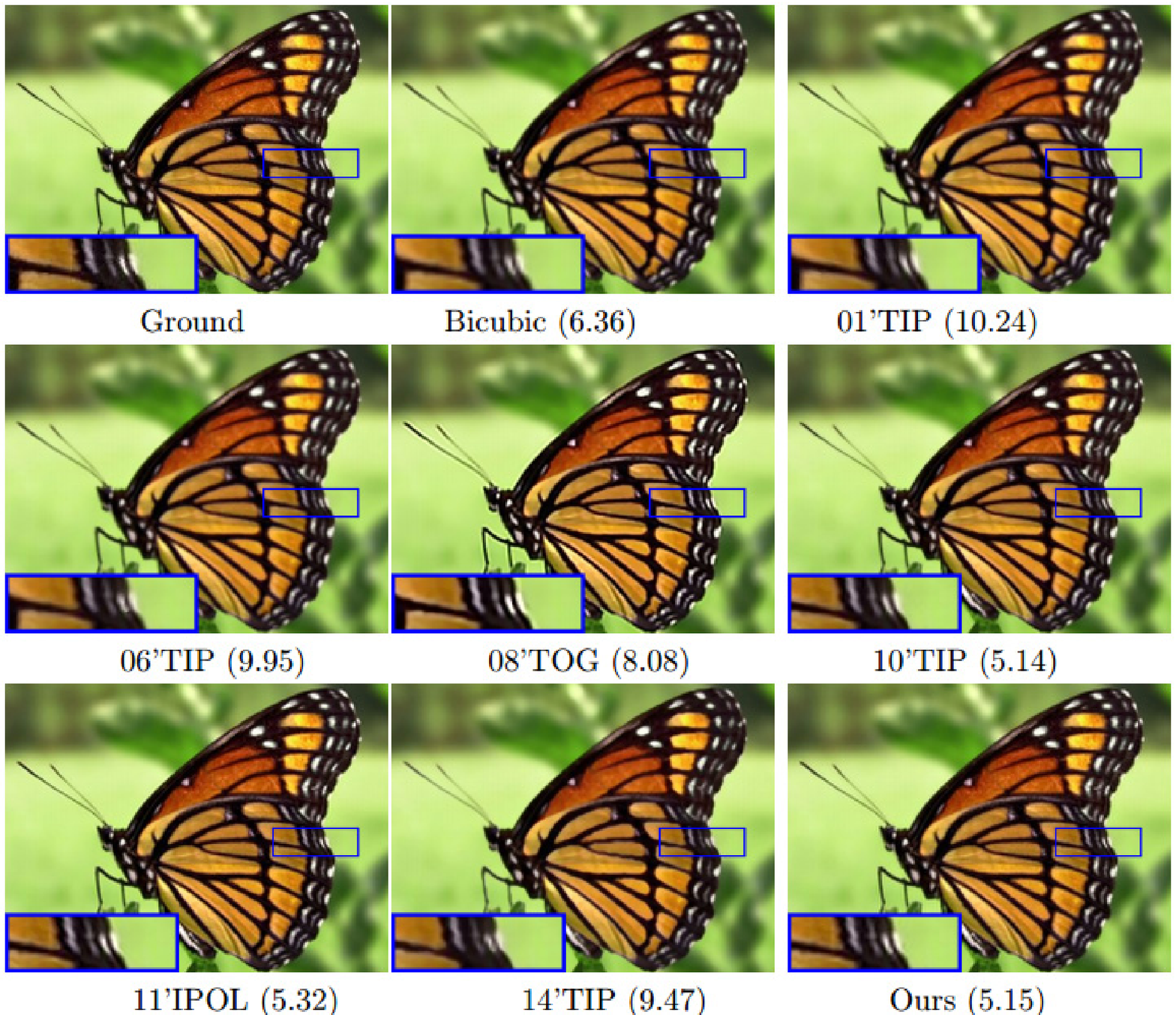}

\end{center}
   \caption{Results of ``butterfly'' by an upscaling factor of 2. First row: ground-truth, results of bicubic interpolation, an edge-directed interpolation ``01'TIP'' \cite{LiOrchard2001}; Second row: results of an edge-guided interpolation ``06'TIP'' \cite{ZhangWu2006}, a fast upsampling method ``08'TOG'' \cite{ShanLi2008}, the learning-based method ``10'TIP'' \cite{YangWright2010}; Third row: results of two interpolation methods ``11'IPOL'' \cite{Getreuer2011} and ``14'TIP'' \cite{WangWu2014},  and the proposed method.}
\label{fig:butterfly}
\end{figure}

In Fig. \ref{fig:butterfly}, we compare the proposed method with some competitive methods. The upscaling factor is 2. The test image ``butterfly'' has obvious smooth background. The results of bicubic interpolation and other two interpolation methods ``01'TIP'', ``06'TIP'' show significant blur effect. Although the result by ``08'TOG'' and ``14'TIP'' keeps sharp edges well, they make image details of non-edge regions oversmoothing. The learning-based method ``10'TIP'' obtains the smallest RMSE and the competitive visual result. The proposed method preserves image details of non-edge regions better and overcomes ring artifacts significantly. In addition, RMSE of our method is almost the smallest one.\\

\noindent\textbf{Parameters selection:} The proposed method involves many parameters, e.g., $\lambda_{1}$, $\lambda_{2}$, $\xi_{1}$, $\xi_{2}$, patch size, etc. However, they are easy to select because the results are not sensitive to the selection of parameters (see $\xi_{1}$, $\xi_{2}$ in Tab. \ref{tab:2} and $\lambda_{1}$, $\lambda_{2}$ in Tab. \ref{tab:3}). Actually, choosing suitable parameters is always a difficulty to many algorithms. Tuning empirically is a popular way for determining parameters. In our work, we obtain the parameters empirically. For instance, for $\lambda_{1}$ and $\lambda_{2}$, we first fix one parameter $\lambda_{1}$, and then tune $\lambda_{2}$ with 10 times change. For instance, we fix  $\lambda_{1} =10^{-2}$, then tune $\lambda_{2}$ by $10^{-3}$, $10^{-4}, 10^{-5}$, $10^{-6}, 10^{-7}$, etc. When we find $\lambda_{2}=10^{-6}$ is the best one for $\lambda_{1}$, then tune $\lambda_{2}$ in the rang of $[10^{-5}, 10^{-7}]$, e.g., $5\times 10^{-6}$ and $5\times 10^{-7}$. Actually, the results around $\lambda=10^{-2}$ and $\lambda_{2}=10^{-6}$ also do not show obvious difference (see Tab. \ref{tab:3}). Fine tuning of parameters might lead to slightly better results. But to make it simple, we set $\lambda_{1} = 10^{-2}$, $\lambda_{2} = 10^{-6}$ in all test examples. Under this setting, the results are already good enough for comparisons. Similarly, we select the other parameters according to this way.\\

\begin{table*}[!htp]
\caption{RMSE around $\xi_{1}=0.1$ and $\xi_{2}=10^{-4}$}\label{tab:2}
\begin{center}
\begin{tabular}{|c||c|c|c|}
\hline
  \backslashbox{$\xi_{2}$ }{$\xi_{1}$}  & $0.8\times 10^{-1}$ & $1\times 10^{-1}$ &  $1.2\times 10^{-1}$ \\
\hline\hline
  $0.8\times 10^{-4}$       & 12.09  & 12.08  &  12.07 \\
\hline
  $1\times 10^{-4}$       & 12.10  & \textbf{12.08}  &  12.07 \\
\hline
  $1.2\times 10^{-4}$     & 12.09  & 12.08  &  12.08 \\
\hline
\end{tabular}
\end{center}
\end{table*}

\begin{table*}[!htp]
\caption{RMSE around $\lambda_{1}=10^{-2}$ and $\lambda_{2}=10^{-6}$}\label{tab:3}
\begin{center}
\begin{tabular}{|c||c|c|c|}
\hline
  \backslashbox{$\lambda_{2}$}{$\lambda_{1}$}  & $0.8\times 10^{-2}$ & $1\times 10^{-2}$ &  $1.2\times 10^{-2}$ \\
\hline\hline
  $0.8\times 10^{-6}$       & 12.09  & 12.08  &  12.08 \\
\hline
  $1\times 10^{-6}$       & 12.09  & \textbf{12.08}  &  12.08 \\
\hline
  $5\times 10^{-6}$     & 12.10  & 12.09  &  12.08 \\
\hline
\end{tabular}
\end{center}
\end{table*}

\begin{figure}
\begin{center}
\includegraphics[width=6in,height=2in]{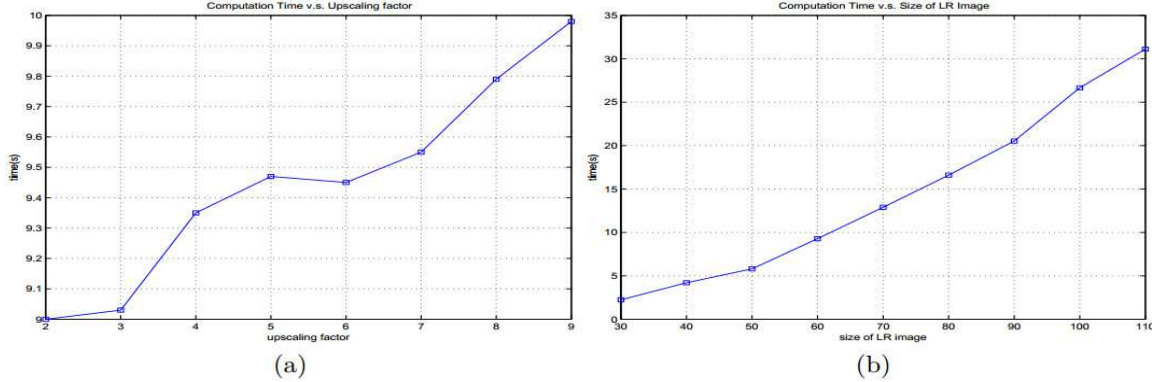}

\end{center}
   \caption{(a) Computation time vs. upscaling factor for low-resolution image with size $60\times 60$; (b)  Computation time vs. size of low-resolution image, the size of low-resolution image is increased from $30\times 40$ to $110\times 110$ and the upscaling factor is always set to be 5. The experimental computer is with 3.47GB RAM, Inter(R) Core(TM) i3-2130 CPU, @3.40GHz.}
\label{fig:time}
\end{figure}

\noindent\textbf{Computation time:} From Fig. \ref{fig:time}(a), we know that the computation time does not increase significantly when upscaling factor is increased from 2 to 9. Using our non-optimized Matlab code, it is below 10 seconds for a low-resolution image $60\times 60$. From Fig. \ref{fig:time}(b), when the size of low-resolution image is from $30\times 30$ to $110\times 110$, the computation time is about increased from 3 seconds to 31 seconds. We can conclude that the computation time mainly depends on the size of low-resolution image but not so significantly on the scaling factor, since the main computation is to compute coefficients $\beta$ via the low-resolution image. Note that, there is a lot of room to speed up the speed. We can use cmex in Matlab to speed up the code that involves a lot of loops. We can also use parallel computing as the computation is done patch by patch.\\

\begin{figure}
\begin{center}
\includegraphics[width=6in,height=4in]{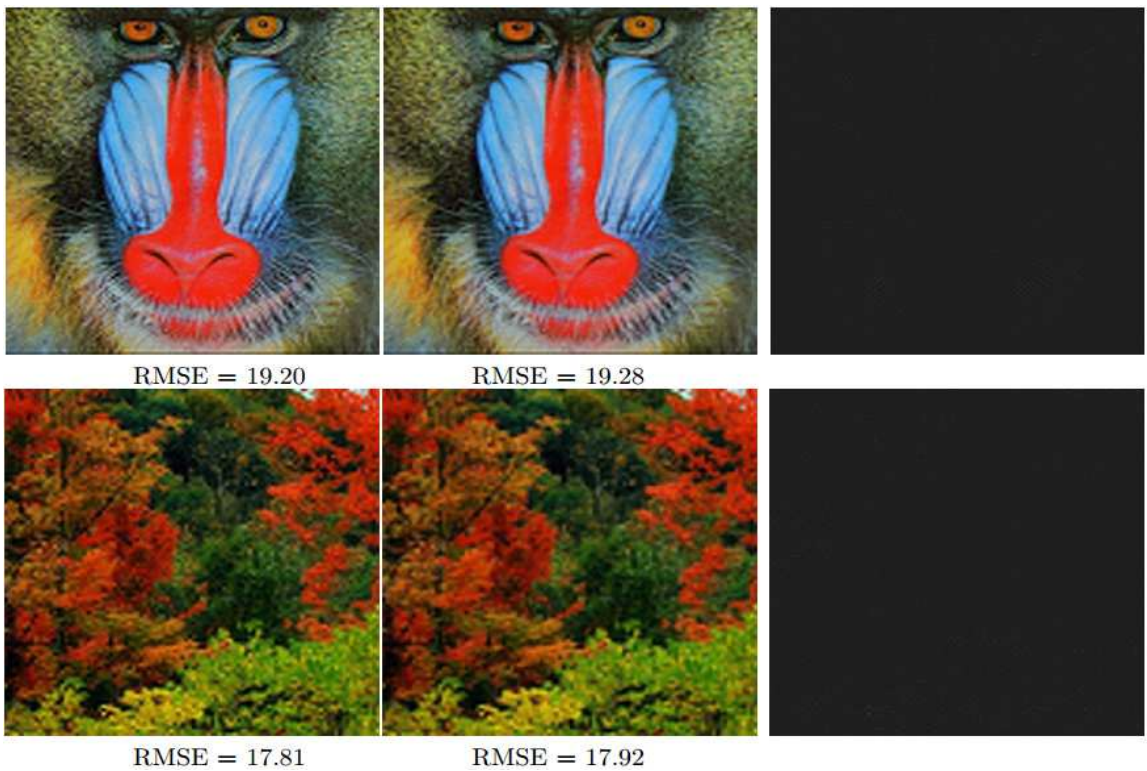}

\end{center}
   \caption{First column: results of model (6) with iterative AHF method; Second column: results of model (6) without iterative AHF method; Third column: error maps (for better vision, we add extra intensity $30/255$ to true error maps). Upscaling factors: 4.}
\label{fig:model6}
\end{figure}

\noindent\textbf{The relation between model (6) and model (6) combined with iterative idea: } It is necessary to illustrate the relation between model (6) and model(6) combine with our iterative idea. Actually, there is only fine difference, especially in image details and edges, between model (6) and model (6) combined with iterative idea. Due to small magnitude of the difference, there is no significant difference between the two methods. In Fig. \ref{fig:model6} we show an example. From Fig. \ref{fig:model6}, it is easy to know that the proposed model (6) combined with the iterative AHF method performs similarly RMSE with the proposed model (6). In addition, the error maps between the two strategies are almost same.

\section{Conclusions}
In this paper, we casted image super-resolution problem as an intensity function estimation problem. We assumed the underlying intensity function, defined on a continuous domain and belonging to a space with redundant basis, could be approximately represented by two classes of approximated Heaviside functions (AHFs).
The representation coefficients were computed by the proposed iterative AHF method using only one low-resolution image. We then applied the coefficients to get high-resolution images. To reduce computation and storage, we applied the proposed iterative AHF method to image patches. For images with smooth backgrounds, we designed a strategy utilizing image gradient to reduce ring artifacts along large scale edges. The method can be applied to any upscaling factors and needs no extra data for learning. In particular, we also discussed the parameter selection and computation time. Many experiments show that the proposed approach outperforms existing competitive methods, both visually and quantitatively.

\section*{Acknowledgement}
\label{sec:Ack}
The first and third authors thank the support by 973 Program (2013CB329404), NSFC (61370147, 61170311), Sichuan Province Sci. \& Tech. Research Project (2012GZX0080). The first author is also supported by NSFC (61402082), Fundamental Research Funds for the Central Universities and Outstanding Doctoral Students Academic Support Program of UESTC.


\end{document}